\documentclass[conference]{IEEEtran}
\usepackage{geometry}
\usepackage{afterpage}

\usepackage{cite}
\usepackage{amsmath,amssymb,amsfonts}
\usepackage{cite}
\usepackage{amsmath,amssymb,amsfonts}
\usepackage{algorithmic}
\usepackage{graphicx}
\usepackage{textcomp}
\usepackage{xcolor}
\usepackage{tcolorbox}
\usepackage{booktabs}
\usepackage{multirow}
\usepackage{makecell}  
\usepackage[normalem]{ulem}
\usepackage{xcolor}
\usepackage{hyperref}
\usepackage[table]{xcolor}
\usepackage{caption}
\usepackage{fontawesome5}

\definecolor{lightgreen}{HTML}{E0F2E9}   
\definecolor{sectiongray}{HTML}{F2F2F2}  

\IEEEoverridecommandlockouts
\begin{document}
\newgeometry{
    top=1.0in,
    bottom=0.75in,
    left=0.75in,
    right=0.75in
}
\afterpage{\restoregeometry}

\title{ProAct-VLM: Pre-Failure Vision-Language Task Replanning with Continuous Perception Feedback}

\author{
Ahmed Nader Ahmed$^{1}$,
Omar Moured$^{2}$,
Mughni Irfan Mohammed Abdul$^{1}$,
Muhayy Ud Din$^{1}$,
Irfan Hussain$^{1}$%
\thanks{$^{1}$Khalifa University, Abu Dhabi, United Arab Emirates.}
\thanks{$^{2}$Sereact GmbH, Stuttgart, Germany.}
}

\maketitle

\begin{abstract}
Long-horizon robotic tasks are vulnerable to unexpected environmental changes that can render planned actions ineffective or unsafe. To address this, robots must detect such changes as they occur, interpret their impact, and adjust their actions accordingly. Traditional rule-based decision-making pipelines are brittle in open-world conditions, as they are hand-tuned for specific scenarios and lack generalization. Vision-Language Models (VLMs) offer a promising alternative as they combine broad world knowledge with unified visual–text reasoning, enabling them to generalize across diverse scenarios and generate accurate, grounded task plans. However, for effective deployment in dynamic real-world settings, VLMs are embedded into frameworks capable of handling uncertainty and environmental changes. Existing frameworks broadly address this reactively, triggering replanning only after execution failures or post-task checks, risking failed actions. Some methods verify conditions before actions, but these discrete checks miss changes occurring during execution. To address this, we present ProAct-VLM, an adaptive, physically grounded task planning framework that integrates VLMs within a real-time perception–feedback loop. ProAct-VLM continuously monitors the environment and re-plans as soon as relevant changes are detected, enabling adaptation before failure occurs. Evaluations against multiple baselines and across different VLM backbones show that our framework improves both success rates and efficiency in dynamic, long-horizon manipulation tasks. Project page available at \href{https://github.com/moured/ProAct-VLM}{https://github.com/moured/ProAct-VLM}.

\end{abstract}


\section{Introduction}



Humans are capable of scene-aware task planning, utilizing environmental context to interpret instructions, formulate appropriate action plans, and adapt those plans in real time in response to changes in the environment \cite{wilensky1983planning}.  Achieving this level of contextual awareness and adaptability for task planning in robotics has long been a challenge. Recent advances in Large Language Models (LLMs) offer a promising step toward this goal. LLMs can function as task planners, interpreting symbolic inputs and natural language commands to generate task plans, thereby reducing reliance on traditional manually engineered modules, offering improved generalization and flexibility\cite{a3}. However, LLMs lack grounding in the physical world, which limits their effectiveness in robotic task planning, a domain that demands an understanding of spatial relationships, robot state, and real-world constraints.

A common approach to address this is to couple LLMs with vision models that analyze the scene and provide structured information, such as object descriptions, poses, and spatial relations, to the LLM \cite{zhao2023chat}. Although this improves grounding, such models often overlook precise, task-dependent details in complex settings, resulting in incomplete or ambiguous world representations. For instance, in a “pour drink into the cup” task, the vision module may detect the cup but fail to determine whether it is empty. Similarly, for “open the door” when an object blocks the path, it may detect both the door and the object without recognizing the object as an obstacle to address. Missing such task-relevant visual cues can result in ineffective or unsafe plans. The emergence of VLMs presents an integrated solution that jointly processes visual and textual inputs within a unified architecture. In robotic task planning, this multimodal capability enables the model to interpret the environment within the context of the task, utilizing visual cues to generate more grounded and executable task plans \cite{hu2023look}.
Although VLMs and LLMs have shown strong capabilities in robotic task planning, many existing approaches operate in offline or static settings \cite{wang2024llm3,wang2024llm}, generating a plan at the start and assuming the environment remains unchanged. This assumption fails in real-world settings, which are inherently dynamic and prone to unexpected changes. Other works that attempt to apply these models in dynamic environments \cite{liu2025robodexvlm,vlm_replan,replan} often incorporate re-planning reactively, triggering it only after execution failure or post-task verification. 

To address this gap, we propose \textbf{ProAct-VLM}, a closed-loop task planning framework that embeds a VLM within a continuous perception–feedback loop. The system maintains persistent environment state awareness during execution and performs \emph{pre-failure adaptation} by re-invoking the planner as soon as task-relevant disturbances (e.g., object addition, removal, or goal modification) are detected. The main contributions of this paper are summarized as follows:

\medskip

\noindent (1) \textbf{Environment State Estimator :}  
A lightweight real-time perception pipeline that detects, tracks, and estimates 6-DoF poses of relevant scene objects, producing a structured environment state and augmented images with persistent object IDs for VLM grounding. This instance-level grounding reduces ambiguity in multi-object scenes and enables more reliable task plan generation compared to other baselines.

\medskip

\noindent (2) \textbf{ProAct-VLM: an adaptive closed-loop task planning framework}: We present a closed-loop integration of VLM reasoning with continuous environment monitoring and instance-level grounding. The framework maintains a structured environment state with persistent object identities, enabling pre-failure replanning through mid-execution disturbance detection. addressing limitations of purely reactive and discrete verification methods.

\medskip

\noindent (3) \textbf{Experimental Validation and Comparative analysis : } 
We designed a dynamic long-horizon manipulation task with scenarios involving object addition, removal, and goal modification to evaluate our framework on a real robot. We further compare our grounding strategy against multiple baselines and conduct an ablation study across VLM backbones to assess their relative strengths in task planning.

\section{Related Work}
Early work such as ViLA \cite{hu2023look} demonstrated the potential of VLMs for task planning by integrating perceptual data directly into the reasoning process to generate  grounded  task plans. Subsequent research embedded VLMs into frameworks for long-horizon task execution in uncertain, dynamic environments, typically triggering re-planning reactively after execution failures. In many cases, a task may be executed correctly at the action level, yet environmental changes can still render the final goal unattainable. For instance, if a robot is instructed to place a cup on a stand and the stand is removed mid-execution, it may still release the cup at the planned coordinates, causing failure. Some methods address such cases by verifying goal completion after task execution, such as Replan \cite{replan} and ReplanVLM \cite{vlm_replan}. Replan employs a VLM for scene grounding and failure reasoning, triggering re-planning only when the motion controller fails to execute an action or when the VLM determines post-execution that the final goal is unmet. Similarly, in ReplanVLM, the VLM receives images of the scene captured before and after task execution and, together with the original plan, determines whether the goal has been achieved; if not, it generates a revised plan. While such methods improve adaptability, they remain reactive, making corrections only after the original plan has been executed. As a result, the robot may still perform unnecessary actions or, in the worst cases, actions that cause damage to objects. Other approaches, whether scene graph-based \cite{rana2023sayplan} or object-centric representation-based \cite{kim2024pre}, follow a similar pattern: they construct a structured representation of the environment (e.g., a scene graph or object-centric state) before each action execution of the task plan, then compare it against an existing reference to detect discrepancies. While effective for pre-action and post-action verification, this discrete checking process means that changes occurring mid-execution of actions often go unnoticed, allowing the agent to proceed with potentially invalid actions. 

An alternative solution is to mimic human behavior by continuously monitoring the environment and updating the plan to preserve long-horizon goals. However, directly using VLMs for real-time change detection, context interpretation, and re-planning remains challenging due to their computational cost and inference latency. To address this, we integrate a lightweight real-time perception pipeline with the VLM in a closed-loop manner in our framework, enabling continuous detection of changes that may invalidate the plan and allowing the framework to react immediately  .

\section{Methods}


\subsection{framework of the proposed solution }
\begin{figure*}[t]
    \centering
    \includegraphics[width=14cm, height=7cm]{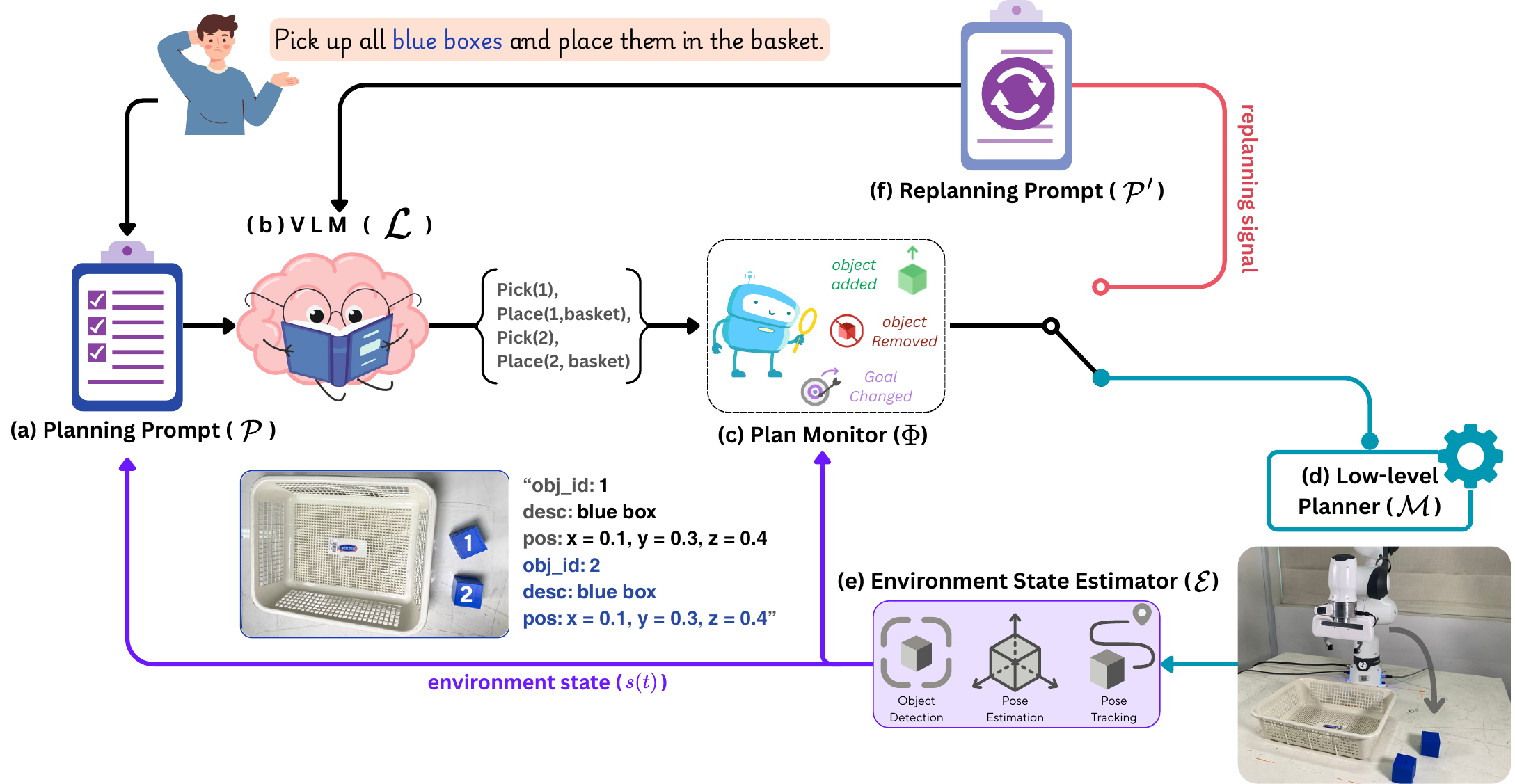}
    \caption{\small Overview of the proposed \textbf{ProAct-VLM} framework. (a) The system begins with a high-level user prompt, which is then converted into a planning prompt. (b) The VLM generates an initial plan based on this input. (e) Meanwhile,the Environment State Estimator continuously constructs a structured scene representation (object IDs, descriptions, poses, and the current Augmented image). (c) The Plan Monitor tracks this state and identifies task-relevant changes (e.g., object addition, removal, or goal modification). When such changes occur, (f) a re-planning prompt is issued to the VLM. (d) The Motion Planner executes the current plan until re-planning is triggered, enabling adaptive task execution in dynamic environments..}

    \label{fig:vlm_replan}
\end{figure*}
The proposed \emph{ProAct-VLM} framework enables    robust and adaptive task execution in dynamic environments by integrating vision-language reasoning with real-time environmental monitoring. As illustrated in Figure~\ref{fig:vlm_replan}, the system consists of four main components: (1) a pre-trained  \emph{VLM} that acts as a black-box task planner generating  symbolic task plans from natural language and visual prompts, (2) a\emph{ Real-time Environment State Estimator} that constructs a structured representation of the scene (including object identities, semantic labels, and 6-DoF poses), (3) a \emph{Plan Monitor} that detects discrepancies in environment states, and (4) a \emph{Low-Level Planner} that translates the symbolic task plan into robot trajectories.

We formulate our task and motion planning framework \emph{ProAct-VLM}, as the following tuple:

\[
\mathcal{F}_{\text{ProAct-VLM}} = \langle\mathcal{P}, \mathcal{E},\mathcal{A},\Phi, \mathcal{L}, \mathcal{M} \rangle
\]

Where:

\begin{itemize}
\item $\mathcal{P}$ denotes the initial \textbf{planning prompt}, formulated in natural language. It specifies the task objective (e.g., ``Pick up all the blue boxes and place them in the basket''). While, $\mathcal{P'}$ represents a \textbf{re-planning prompt}, which may include a modified task objective along with additional textual cues indicating that the current scenario requires re-planning.

  \item $\mathcal{E}$ is the \textbf{Environment State Estimator}, which maps visual observations to the state of the environment $s(t)$: 
  \begin{equation}
  \mathcal{E}: \text{RGB-D image} \rightarrow s(t) \in \mathcal{S}
  \end{equation}
  we assume that \text{RGB-D image} captures the entire workspace of the robot.
  $s(t)$ is the environment state at time $t$ which includes object IDs, semantic labels, 6-DoF poses, and an augmented image of the environment annotated with object IDs, as illustrated in Figure \ref{fig:vlm_replan} 
  
  \item $\mathcal{A}$ is the set of \textbf{primitive actions}, where each action $a(o)$ acts on an object referenced by an ID $o \in O$. A feasible action is indicated by $a(\tau)$ if the motion planner finds a valid trajectory $\tau$.
    \item $\Phi$ is the \textbf{plan monitor}, modeled as a binary function that outputs $1$ to trigger replanning and $0$ otherwise. Replanning is initiated when one or more disturbances are detected, defined as \[     \Delta \subseteq \{\text{Object Added}, \text{Object Removed}, \text{Goal Changed}\}     \] 
    
    
    These disturbances are inferred by comparing the current state $s(t)$ with the previous state $s({t-\Delta t})$, and checking if the user gives a new task objective, The decision rule for triggering re-planning is defined mathematically as:
    
    \begin{equation}
        \Phi(s(t), s({t-\Delta t})) =
        \begin{cases}
            1, & \text{if } \Delta \neq \emptyset \\
            0, & \text{otherwise}
        \end{cases}
        \label{eq:plan_monitor}
    \end{equation}

  
  \item \noindent
    $\mathcal{L}$ is the (VLM) model which acts as the task planner and is defined as :
    \begin{equation}
    \mathcal{L} : \mathcal{P} \times s(t=0) \times \mathcal{A} \rightarrow a
    \end{equation}
    \end{itemize}

As illustrated in Figure \ref{fig:vlm_replan}. Initially, at  $t=0$, the VLM model receives the prompt $\mathcal{P}$, which is augmented with the initial environment state $s(t = 0)$ and the set of available primitive actions $\mathcal{A}$ (prompt format shown in Figure~\ref{fig:prompt_template}). It then outputs the task plan as a set of actions $a \subset \mathcal{A}$. The generated action sequence is defined as:
\[
a = \{ a_0(o_0), a_1(o_1), \dots, a_t(o_i) \}
\]

for each action $a_t(o_i)$, the \textbf{low-level planner} $\mathcal{M}$  resolves the object ID $o_i$ to its pose from $s(t)$ and uses a motion planner to compute a valid trajectory $\tau_t$. The resulting  feasible action  $a_t(\tau_t)$ is then executed, moving the robot from its current state toward the desired goal state.

During the execution of $a$, $\mathcal{E}$ continuously updates $s(t)$ in real time. This state is monitored by the plan monitor $\Phi$ to detect any disturbances $\Delta$ that may require re-planning. When such a disturbance is detected, execution stops and a new prompt $\mathcal{P'}$ is generated, augmented with the current state $s(t)$, and passed to the VLM to generate an updated action sequence:

\begin{equation}
    \mathcal{L} : \mathcal{P'} \times s(t) \times \mathcal{A} \rightarrow a
\end{equation}

This process continues iteratively until the goal condition is satisfied that is, the final plan $a$ is successfully executed without interruption. this ensures system adaptability and robustness in dynamic environments

\subsection{Environment state estimator $\mathcal{E}$ }\label{AA}

\begin{figure*}[t]
    \centering
    \includegraphics[width=14cm, height=7cm]{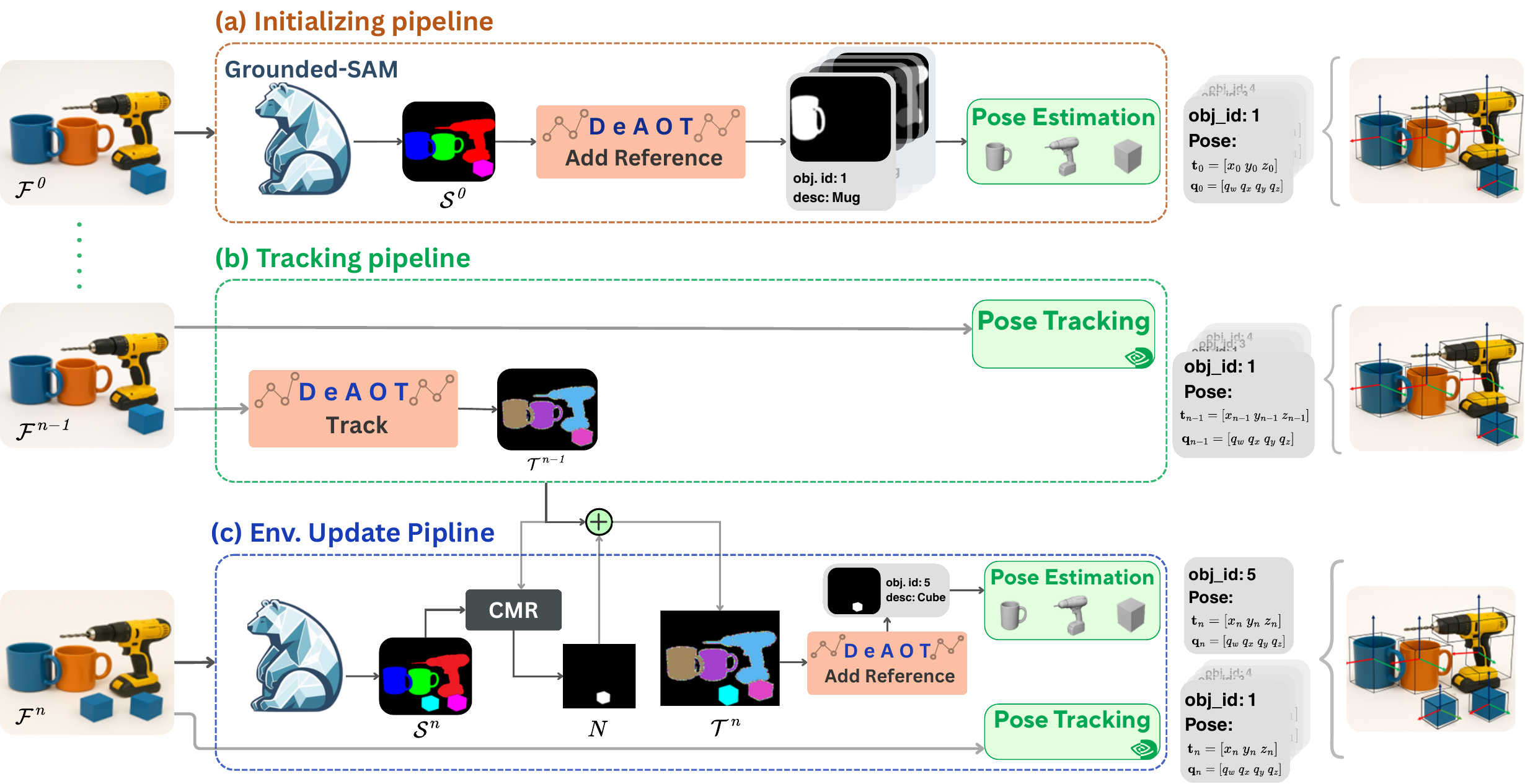}
    \caption{\small Overview of the proposed perception pipeline. The system operates in three cyclic phases: Initialization (a), Tracking (b), and Environment Updating (c). In (a) Grounded-SAM detects objects from an input image using text prompts, and generates labeled segmentation masks. These masks are linked to CAD models and processed by FoundationPose for initial 6-DoF pose estimation. During (b), DeAOT tracks object masks across frames, while FoundationPose perform pose tracking for already registered objects. In (c), object detection is re-invoked periodically to identify new objects using a Comparing Mask Results (CMR) strategy, followed by pose estimation and reference update.}
    \label{fig:env_state_monitor}
\end{figure*}

The \textbf{Environment State Estimator} is built upon three key components:

\noindent (1) \textbf{Grounded-SAM} is a vision-language pipeline that combines \textbf{Grounding DINO} \cite{liu2024grounding}  and the \textbf{Segment Anything Model (SAM)} \cite{kirillov2023segment} to perform open-vocabulary object detection and segmentation from natural language prompts. Grounding DINO detects objects by aligning textual and visual features to produce bounding boxes, while SAM uses these boxes to generate segmentation masks, enabling accurate zero-shot  object detection and segmentation.

\noindent (2) 
\textbf{DeAOT} \cite{yang2022decoupling} is a state-of-the-art video object segmentation model used for accurate multi-object tracking. Its ability to maintain spatio-temporal consistency across frames, even under occlusion and fast motion, is critical for reliable environment monitoring.


\noindent (3) 
\textbf{FoundationPose} \cite{wen2024foundationpose} is a unified framework for 6-DoF object pose estimation and tracking in both model-based and model-free settings. We adopt the model-based variant, which leverages CAD models to achieve accurate pose estimation for manipulation tasks such as pick-and-place and assembly.


These components are integrated into the perception pipeline illustrated in Figure~\ref{fig:env_state_monitor}, designed to estimate $s(t)$. To formally describe the system, we define the following variables.

\begin{itemize}
    \item $F^{n} \in {R}^{H \times W}$: $n^{th}$ RGB frame captured by the camera 
    \item $S^{n} \in {R}^{H \times W}$: Grounded-SAM generated segmentation masks of the $n^{th}$ RGB frame
    \item $T^{n} \in {R}^{H \times W}$:  Tracked masks generated by the  DeAOT tracker of the $n^{th}$ RGB frame
    \item $N \in{R}^{H \times W}$: Masks representing newly detected objects,
    \item $t \in (0, 1)$: A predefined threshold to determine object novelty.
    \item $n$:                                                                   A fixed number of frames, where at each $kn$ frame environment update will occur.
\end{itemize}
Here $H$ and $W$ denote the height and width of the image dimensions, respectively.
The system operates in a cyclic perception loop composed of three primary phases: Initialization, Tracking, and Environment Update.

At the beginning, the system enters the \textbf{Initialization phase}, whose primary objective is to bootstrap the perception pipeline with an initial object detection, segmentation, and 6-DoF pose estimation.  First, the initial camera frame $F^0$ is processed by the Grounded-SAM pipeline using a prompt listing predetermined relevant objects (e.g., mug, drill, cube). The pipeline outputs segmentation masks $S^0$ for each object identified in the prompt. Each mask is associated with its corresponding object label and then assigned a unique identity for subsequent tracking. These labeled masks are stored as references for the DeAOT tracker, which uses them to track objects across subsequent frames. Each labeled mask is matched to its corresponding CAD model from a preloaded library, These paired inputs are then passed to the FoundationPose module to perform 6-DoF pose estimation. 
In addition to estimating poses, FoundationPose registers visual and geometric features for future tracking. By the end of the initialization phase, the system obtains the intial structured  scene representation of the environment $s(0)$.
Immediately after processing the initial frame, the system transitions into the \textbf{Tracking phase}, where the DeAOT tracks detected objects across subsequent frames, producing tracked object masks denoted as $T^{n-1}$ in Figure~\ref{fig:env_state_monitor} Simultaneously, FoundationPose tracks the poses of previously registered objects. Together, these modules update $s(t)$.

After a certain number of frames $n$ the system enters the \textbf{Environment Updating phase} to check for newly introduced objects. 
 object detection is re-invoked for the $n^{th}$ frame $F^{n}$ using Grounded-SAM with the same predetermined text prompts producing segmentation masks $S^{n}$ as shown in Figure~\ref{fig:env_state_monitor}, For the previous frame $F^{n-1}$ the DeAOT tracker would have produced the tracked masks $T^{n-1}$ which is used in the process of detecting newly introduced objects, following the approach proposed in \cite{kirillov2023segment} where the masks representing candidate new objects, denoted as  $N\_cand$ is computed by taking the pixel-wise product of $S^{n}$ and the binary background mask from the previous tracking result $T_0^{n-1}$ :
\begin{equation}
N\_cand = T_0^{n-1} \odot S^{n}
\label{eq:new_object_mask}
\end{equation}

For a candidate object $x$ , let $x_N$ and $x_s$ denote the number of pixels occupied by $x$ in $N\_cand$  and $S^{n}$ respectively. We define the Comparing Mask Results (CMR) indicator function as:

\begin{equation}
\text{CMR}(x) =
\begin{cases}
1, & \text{if } \dfrac{x_N}{x_s} > t \\
0, & \text{otherwise}
\end{cases}
\label{eq:cmr}
\end{equation}

An object is classified as \emph{new} if a significant portion of it lies in the background region according to threshold $t$,In our implementation
$t$ is empirically set to 0.5 based on validation experiments. The final new objects masks $N$ is obtained by applying the CMR filter to all candidates in $N_{\text{cand}}$, all new masks in $N$ are then assigned unique IDs, these new masks are merged with the previously tracked masks $T^{n-1}$ to produce the updated tracked mask $T^{n}$, as follows:
\begin{equation}
T^{n} = T^{n-1} + N
\label{eq:new_track_mask}
\end{equation}

\begin{figure}[t]
    \centering    \includegraphics[width=8cm, height= 5cm]{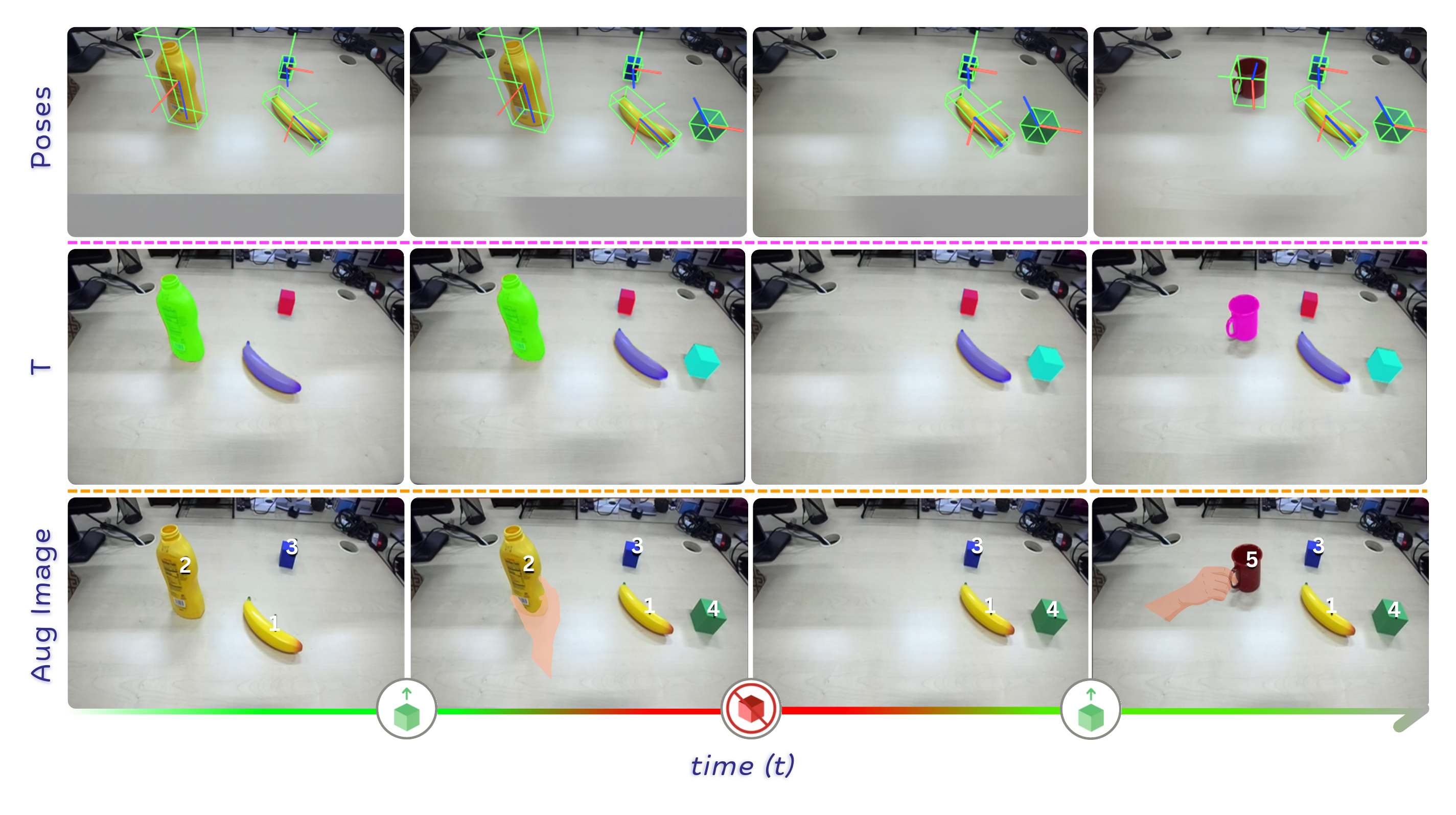}
    \caption{\small illustrates of $\mathcal{E}$  in action. The framework detects, segments, and estimates object poses as the scene evolves, correctly handling object additions and removals.  }
    \label{fig:monitor_test}
\end{figure}

Which is then stored as a new reference for DeAOT tracking. For all newly detected objects, Foundationpose estimates their 6-DoF pose. At the same time, continuing to track the pose of previously detected objects. The tracking–environment update cycle repeats every $Kn$ frames until the system is terminated. Figure \ref{fig:monitor_test} demonstrates an example of $\mathcal{E}$ in action, where the system detects object additions and removals and updates their poses accordingly. 
The resulting augmented image with overlaid object IDs enables instance-level grounding for the VLM.

We measured the end-to-end runtime of $\mathcal{E}$  on the workstation used for all experiments (NVIDIA RTX 4090 GPU, 64\,GB RAM, Ubuntu 22.04). During normal operation, the tracking phase runs at approximately 30\,Hz (33\,ms per frame). 
Periodic full environment updates, which re-invoke detection and pose initialization, take approximately 200 \, ms (5 \, Hz).  Since full updates are invoked intermittently while tracking runs continuously, the effective monitoring frequency remains above 20\,Hz in practice.  Given that primitive manipulation actions last several seconds, this frequency is sufficient to detect disturbances and update the environment state before the action is completed.

\begin{figure}[t]
    \centering
    \includegraphics[width=\linewidth,height=4cm]{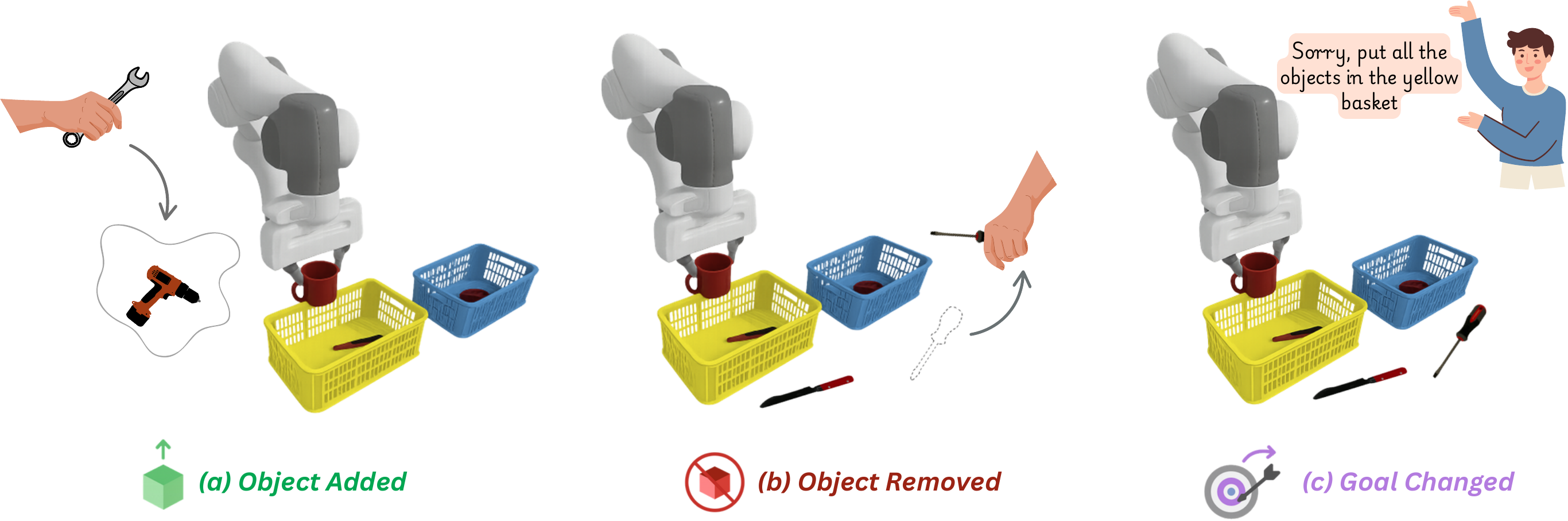}
    \caption{\small Disturbance types used in experiments: (a) object addition, (b) object removal, (c) goal change.}
    \label{fig:types_of_disturbances}
\end{figure}





\section{Results}

\subsection{Experiment Domain and Evaluation Metrics}

To evaluate ProAct-VLM, we designed a real-world sorting task in which a Franka robotic arm is instructed to sort objects into two distinct baskets: one for shape items and the other for food items. The experiment is conducted in a real-world setting using objects from the YCB dataset. The robot operates using two primitive actions that form the basis of any task plan:

\begin{itemize}
    \item \texttt{pick(obj\_id)}: pick up \texttt{obj\_id}  
    \item\texttt{place(obj\_id,basket\_id)}: place \texttt{obj\_id} in \texttt{basket\_id} 

\end{itemize}

We chose sorting as our experimental task as it encapsulates the core challenges of real-world task planning while remaining well-defined and easy to evaluate. As a long-horizon manipulation problem, sorting requires executing a sequence of actions, making it naturally susceptible to environmental changes and disturbances over time. It also demands semantic reasoning to recognize and categorize objects, as well as spatial relationship reasoning to understand the current state of the environment in the context of the task (e.g., which object is in which basket). Moreover, it is an intuitive task highly relevant to real-world applications. A core objective of our framework is to operate reliably in dynamic scenarios. To evaluate this, we introduce disturbances during the sorting task execution that match those detected by the plan monitor illustrated in Fig. \ref{fig:types_of_disturbances}. Using these disturbances, we define seven distinct test scenarios, each incorporating one or more disturbances introduced at different times during task plan execution. The details of these scenarios are explained in Table \ref{tab:scenario_definitions}.  This setup allows us to simulate diverse and unpredictable situations that may arise during task execution.

To evaluate performance, we use two primary metrics. The first is \textbf{success rate}, which measures whether the task plan generated by the VLM successfully achieves the intended goal. The second is \textbf{efficiency}, defined as:

\begin{equation}
\text{Efficiency} = \left( \frac{\text{ Minimum \# of Steps to Goal}}{\text{\# of Steps in VLM Plan}} \right) \times 100%
\end{equation}
This metric is computed only when the VLM-generated plan successfully achieves the goal. Here, Minimum Number of Steps to Goal denotes the length of the optimal (shortest) plan. while the Number of Steps in the VLM Plan refers to all executable primitive actions in the plan, including redundant or unnecessary ones. Thus, this metric quantifies how efficiently the VLM-generated plan achieves the task.




\definecolor{lightblue}{RGB}{220, 240, 255}
\definecolor{lightorange}{RGB}{255, 204, 153} 
\definecolor{lightgreen}{RGB}{220, 255, 220}
\definecolor{darkgreen}{RGB}{0,100,0}

\subsection{Ablation study}
 $\mathcal{E}$ generates an augmented image with overlaid object IDs and provides class labels for each object. This information forms the VLM input prompt (Fig.~\ref{fig:prompt_template}), which is further augmented with additional instructions during re-planning. This prompt formulation is used in each scenario as the input to the VLMs. Based on this setup, we conduct an ablation study across the seven previously defined scenarios (Table~\ref{tab:scenario_definitions}), evaluating five models, three proprietary (GPT-4o, Gemini 2.5 Pro, Claude Opus 4) and two open-source (Qwen 2.5-VL, LLaMA Maverick) based on success rate and efficiency, averaged over 10 trials per scenario. Results in Table~\ref{tab:llm_ablation_subcolumns} show that Gemini 2.5 Pro achieved perfect performance across all scenarios, followed by GPT-4o with strong averages of 89\%  success and 93\%  efficiency. In contrast, Claude Opus 4 underperformed significantly. Among open-source models, LLaMA Maverick outperformed Qwen 2.5-VL, indicating relatively better spatial reasoning. Overall, Gemini 2.5 Pro demonstrated the most consistent performance, suggesting that its architecture is particularly effective for spatial reasoning and long-horizon task planning in dynamic environments.

\subsection{Comparison to baselines(Grounding)}
\begin{figure}[t]
\centering
\begin{tcolorbox}[width=0.5\textwidth, colback=gray!5, colframe=black, boxrule=0.2pt, fonttitle=\bfseries]
\footnotesize 
\noindent
\begin{minipage}[t]{0.49\textwidth}
\begin{tcolorbox}[colback=lightblue, colframe=lightblue, boxrule=0.2pt, sharp corners=south]
\centering
\includegraphics[width=\linewidth,height=2cm]{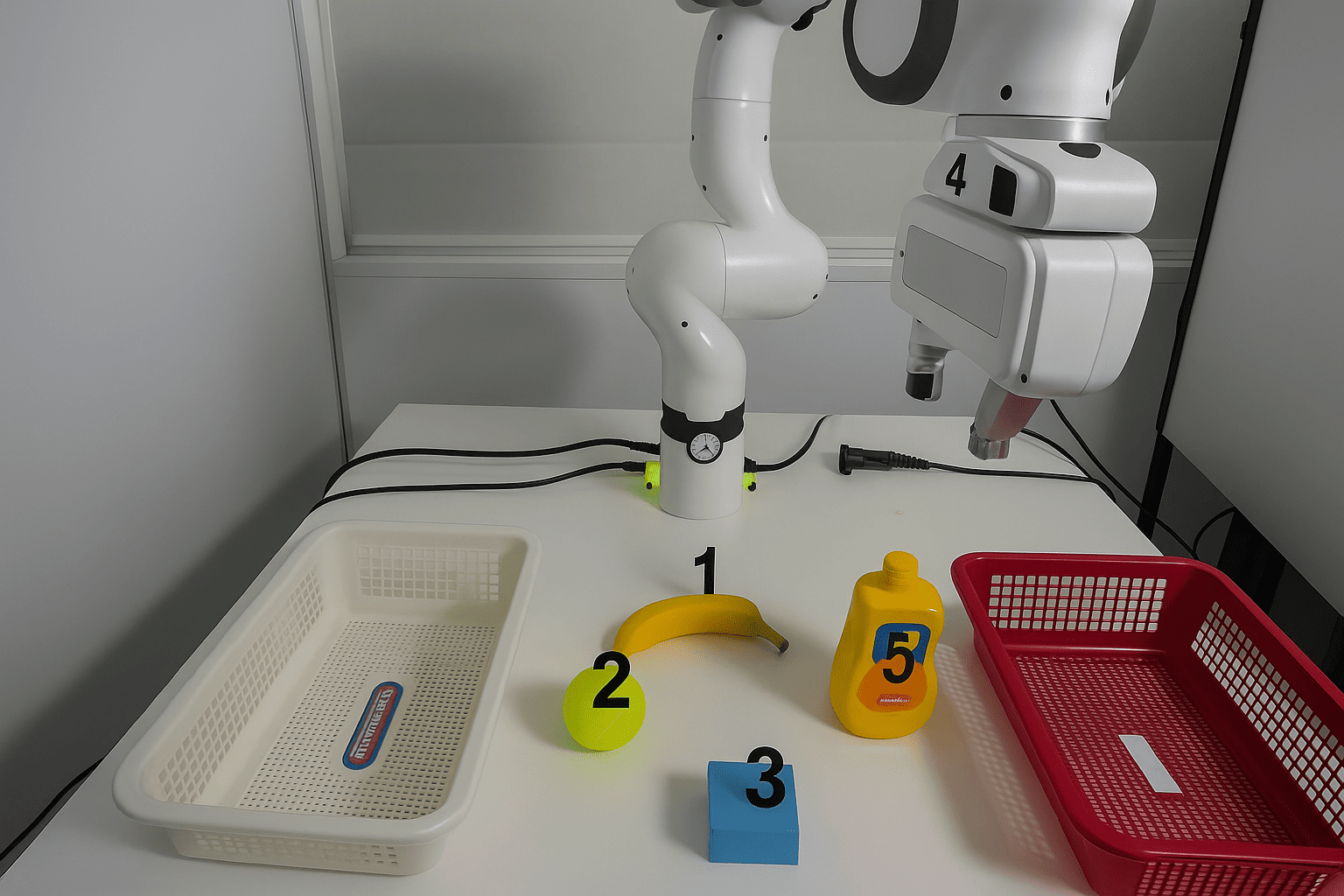}
\end{tcolorbox}
\end{minipage}%
\hfill
\begin{minipage}[t]{0.49\textwidth}
\begin{tcolorbox}[colback=lightorange, colframe=lightorange, boxrule=0.2pt, sharp corners=south]
\centering
\includegraphics[width=\linewidth,height=2cm]{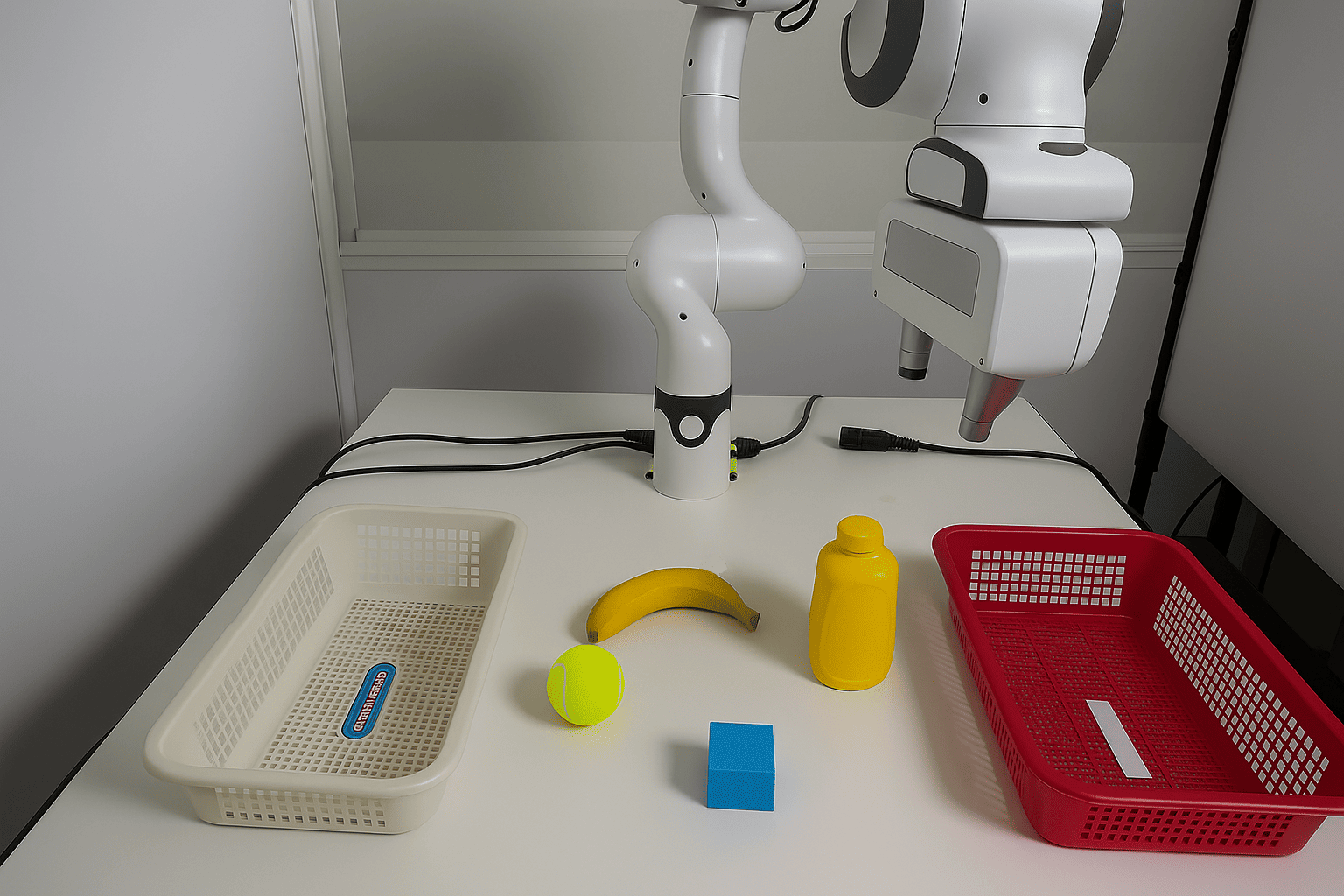}
\end{tcolorbox}
\end{minipage}

\textbf{System message:} You are an AI robot that generate a plan of actions to reach the goal, we have an environment that has a robot arm, two baskets a  red\_basket and a white\_basket and several objects, \textcolor{red}{you need to re-plan because something happened in the environment which requires a new plan, please be efficient}
\par\vspace{\baselineskip}
\textbf{Env\_state(t)} : There are several objects in the environment with these ids : 1,2,3,5
\begin{itemize}
  \item The object with id 1 is a tennis ball, the object with id 2 is a banana............................................
\end{itemize}
\par\vspace{\baselineskip}
\textbf{Object\_poses(t):} 
\begin{itemize}
  \item object with id 1 is at position [-0.25, -0.25, 0.02]...........
  \item The red\_basket occupies a rectangular area defined by x-range from x\_min=0.329 to x\_max=0.59, y-range from y\_min=0.37 to y\_max=0.6....The white\_basket occupies...
\end{itemize}
\par\vspace{\baselineskip}
\textbf{Actions available:} 
Robot has these primitive actions:
\begin{itemize}
  \item \texttt{pick([obj\_id], \{\})}
  \item \texttt{place([obj\_id], \{basket\_id\})}
\end{itemize}
\par\vspace{\baselineskip}
\textbf{Goal:} Sort out objects to shape or kitchen items.  
\par\vspace{\baselineskip}
\textbf{Output format:} Please generate output step-by-step. Output format (JSON):\\
\texttt{\{"Full Plan": ["pick(['obj\_id'], \{\})", "place(['obj\_id'], \{'basket\_id'\})", ... ] \}}
\end{tcolorbox}

\caption{\small Prompt templates used by \textbf{VLM-Replan}. The components in orange are specific to the LLM3+Image version, while the blue are specific to our Augmented version, \textcolor{red}{the red text} represents what we add to make the prompt for re-planning.}
\label{fig:prompt_template}
\end{figure}

Our grounding strategy leverages an augmented image with overlaid object IDs together with a text specifying the class of each ID, all generated by $\mathcal{E}$, and includes them in the VLM prompt. The complete prompt structure is illustrated in Figure~\ref{fig:prompt_template}. We evaluate this strategy against several relevant baselines. One baseline is based on the LLM3 \cite{wang2024llm3} approach, a text-based method that incorporates object positions, as well as the bounding box coordinates of each basket, into the prompt. However, providing only text-based spatial information restricts the baseline, since multi-modal models are designed to leverage both visual and textual cues. By extending this baseline to include a raw image of the scene, we enable the VLM to utilize both modalities, resulting in a more appropriate and fair baseline for comparison. Other VLM task-planning baselines, such as VILA \cite{hu2023look} and Replan VLM \cite{vlm_replan} rely solely on the scene image to generate a plan based on the object class labels (e.g., (pick(blue\_cube))), then use other vision modules to detect this class and ground it in the scene, However, these approaches suffer from ambiguity when multiple objects share the same class (e.g., two identical blue cubes) lacking instance-level grounding compared to our approach. A closely related baseline to ours is introduced in Open-World Grasper(OWG)\cite{tziafas2024openworldgraspinglargevisionlanguage}, which also employs an augmented image with overlaid object IDs. However, a key distinction lies in how object information is handled: in their approach, the VLM is used to infer object categories directly from the augmented image. In contrast, our method explicitly provides the class label associated with each object ID through the Env\_state(t) part of the prompt. Figure~\ref{fig:prompt_template} also shows the LLM3+Image prompt by omitting the blue-highlighted parts, which are relevant to our baseline.  
The RePlan-VLM/VILA prompt is obtained by removing Object\_poses(t) from the LLM3+Image prompt,  
while the OWG prompt is obtained by removing Object\_poses(t) from our baseline prompt. We evaluated the top three VLMs from our ablation study (GPT-4o, Gemini 2.5 Pro, and LLaMA 4 Maverick) across the seven disturbance scenarios using the baselines mentioned earlier. 
The results are summarized in Table~\ref{tab:prompt_per_scenario_compact}. 
The results highlight the clear advantage of our grounding approach, with our prompt consistently outperforming all baselines. The key difference between our approach and the LLM3+Image baseline lies in the type of instance-level grounding provided. While LLM3+Image includes an image, it grounds instances through positional data. For example, in scenarios with multiple objects of the same class (e.g., several cubes), the VLM must rely on coordinates to distinguish between them, such as deciding which cube is inside or outside a basket. By contrast, our approach provides explicit visual instance-level grounding, reducing such ambiguity. The improved performance we observe suggests that VLMs capture spatial relationships more effectively through visual grounding than through structured positional data alone.
Our approach outperformed the OWG baseline, despite their structural similarity. The key difference lies in how object categories are obtained: OWG relies on the VLM to infer classes directly from the augmented image, whereas our approach uses $\mathcal{E}$ to assign class labels to each object. This distinction is essential, as specialized vision models have been shown to surpass general-purpose VLMs in standard detection tasks \cite{ramachandran2025doesgpt4ounderstandvision}. Approximately 30 \% of OWG failures result from misclassification (e.g., confusing a cube with a sugar box), which directly results in incorrect task plans.
Similarly, VILA/RePlan-VLM performs poorly in scenarios that require instance-level grounding, such as Scenarios 2 and 7, which involve two identical blue boxes. In such cases, the VLM may output ambiguous actions (e.g., pick(blue\_box)) or overly specific relational instructions (e.g., pick(blue\_box\_on\_the\_table)). However, grounding models like Grounding Dino or YOLO are not designed to interpret relational prompts and, therefore, fail to resolve the correct instance.

\begin{table}[!t]
\centering
\caption{\small Defined disturbance scenarios used in our evaluation.}
\label{tab:scenario_definitions}
\setlength{\tabcolsep}{3pt}            
\renewcommand{\arraystretch}{1.1}      

\begin{tabular}{@{}l
    >{\raggedright\arraybackslash\footnotesize\ttfamily}p{3.6cm}
    >{\raggedright\arraybackslash\small}p{3.9cm}@{}}
\toprule
\textbf{Sc.} & \textbf{$Env\_state(t)$} & \textbf{Description} \\
\midrule
S1 & Red\_Basket: empty; White\_Basket: empty; Robot\_gripper: empty;

table: cube, banana, tennis ball, mustard bottle & Start sorting; no objects sorted yet. \\
\cmidrule(lr){1-3}
S2 & Red\_Basket: cube; White\_Basket: empty; Robot\_gripper: empty; 

table: banana, tennis ball, mustard bottle, \textcolor{darkgreen}{cube}
   & Extra object added mid-sorting. \textcolor{darkgreen}{(a)}\\
\cmidrule(lr){1-3}
S3 & Red\_Basket: cube, tennis ball, \textcolor{darkgreen}{baseball};

White\_Basket: banana; Robot\_gripper: empty;

table: mustard bottle
   & Object added to the correct basket mid-sorting. \textcolor{darkgreen}{(a)} \\
\cmidrule(lr){1-3}
S4 & Red\_Basket: cube, tennis ball; White\_Basket: banana, \textcolor{darkgreen}{baseball}; 

Robot\_gripper: empty; 

table: mustard bottle
   & Object added to the wrong basket mid-sorting. \textcolor{darkgreen}{(a)}\\
\cmidrule(lr){1-3}
S5 & Red\_Basket: cube; White\_Basket: empty; Robot\_gripper: banana; 

table: \textcolor{red}{\sout{tennis ball}}, mustard bottle
   & One object removed. \textcolor{red}{(b)} \\
\cmidrule(lr){1-3}
S6 & Red\_Basket: cube; White\_Basket: empty; Robot\_gripper: empty; 

table: banana, tennis ball, mustard bottle
   & New instruction; goal changed. \textcolor{purple}{(c)} \\
\cmidrule(lr){1-3}
S7 & Red\_Basket: cube, tennis ball; White\_Basket: banana; Robot\_gripper: empty; 

table: mustard bottle, \textcolor{darkgreen}{cube}
   & Extra object added; goal changes near completion.
   
   \textcolor{darkgreen}{(a)} +  \textcolor{purple}{(c)} \\
\bottomrule
\end{tabular}
\end{table}

\begin{table}[!t]
\centering
\caption{\small Results of Ablation Study}
\label{tab:llm_ablation_subcolumns}
\renewcommand{\arraystretch}{1.1}
\setlength{\tabcolsep}{2.5pt}
\scriptsize
\begin{tabular}{l
                cc cc cc cc cc}
\toprule
\textbf{Scenario} 
& \multicolumn{2}{c}{\textbf{GPT-4o}} 
& \multicolumn{2}{c}{\textbf{Gemini 2.5}} 
& \multicolumn{2}{c}{\textbf{Claude 4}} 
& \multicolumn{2}{c}{\textbf{Qwen 2.5}} 
& \multicolumn{2}{c}{\textbf{Llama 4}} \\
\cmidrule(lr){2-3}
\cmidrule(lr){4-5}
\cmidrule(lr){6-7}
\cmidrule(lr){8-9}
\cmidrule(lr){10-11}
& Succ & Eff & Succ & Eff & Succ & Eff & Succ & Eff & Succ & Eff \\
\midrule
S1 & 100 & 100 & 100 & 100 & 100 & 100 & 20  & 100 & 100 & 100 \\
S2 & 100 & 100 & 100 & 100 &  40 & 100 & 70  & 100 &  60 &  97 \\
S3 &  80 &  80 & 100 & 100 &   0 &   0 & 60  &  70 & 100 & 100 \\
S4 &  80 & 92.5& 100 & 100 &   0 &   0 & 40  &  80 &  10 &  50 \\
S5 & 100 & 96.6& 100 & 100 &   0 &   0 & 50  &  90 & 100 &  96 \\
S6 &  90 &  95 & 100 & 100 &   0 &   0 & 40  &  75 &  90 & 100 \\
S7 &  70 &  84 & 100 & 100 &  10 &  66 &  0  &   0 &  30 &  83 \\
\midrule
\textbf{Avg.} 
   & 89  & 93  & \cellcolor{lightgreen}\textbf{100} & \cellcolor{lightgreen}\textbf{100} 
   & 21.4 & 38 & 40 & 73.6 & 70 & 90 \\
\bottomrule
\end{tabular}
\end{table}






\subsection{Comparison to Reactive Replanning}

\begin{table}[t]
  \centering
  \caption{\small Results of Comparative Studies .}
  \label{tab:prompt_per_scenario_compact}
  \small
  \resizebox{\columnwidth}{!}{%
    \arrayrulecolor{gray!50}%
    \begin{tabular}{@{}llccccccc|c@{}}
      \toprule
      \textbf{Model} & \textbf{Prompt} & \textbf{S1} & \textbf{S2} & \textbf{S3} & \textbf{S4} & \textbf{S5} & \textbf{S6} & \textbf{S7} & \textbf{Avg} \\
      \midrule
      \multirow{4}{*}{{GPT-4o}} 
        & \colorbox{lightorange}{LLM3+Image}  & 90 & 40 & 80 & 30 & 100 & 60 & 30 & 61.4 \\
        & \colorbox{gray!20}{OWG} & 100 & 80 & 70 & 70 & 100 & 90 & 70 & 83 \\
        & \colorbox{lightblue}{Ours}     & \textbf{100} & \textbf{100} & \textbf{80} & \textbf{80} & \textbf{100} & \textbf{90} & \textbf{70} & \textbf{89} \\
        & \colorbox{yellow!30}{VILA/REPLAN-VLM} & 100 & 0 & 100 & 40 & 80 & 30 & 0 & 50 \\
      \midrule
      \multirow{4}{*}{{Gemini 2.5 Pro}} 
        & \colorbox{lightorange}{LLM3+Image}  & 100 & 90 & 100 & 100 & 100 & 100 & 100 & 98.6 \\
        & \colorbox{gray!20}{OWG} & 100 & 100 & 90 & 100 & 100 & 100 & 100 & 98.6 \\
        & \colorbox{lightblue}{Ours}    & \textbf{100} & \textbf{100} & \textbf{100} & \textbf{100} & \textbf{100} & \textbf{100} & \textbf{100} & \textbf{100} \\
        & \colorbox{yellow!30}{VILA/REPLAN-VLM} & 100 & 0 & 100 & 100 & 100 & 100 & 0 & 71.4 \\
      \midrule
      \multirow{4}{*}{{Llama 4 Maverick}} 
        & \colorbox{lightorange}{LLM3+Image} & 90 & 0 & 100 & 50 & 50 & 60 & 10 & 51.4 \\
        & \colorbox{gray!20}{OWG} & 80 & 50 & 90 & 20 & 50 & 30 & 10 & 47.1 \\
        & \colorbox{lightblue}{Ours}    & \textbf{100} & \textbf{60} & \textbf{100} & \textbf{10} & \textbf{100} & \textbf{90} & \textbf{30} & \textbf{70} \\
        & \colorbox{yellow!30}{VILA/REPLAN-VLM} & 40 & 0 & 90 & 50 & 30 & 0 & 0 & 52.5 \\
      \bottomrule
    \end{tabular}%
  }
\end{table}

We compare ProAct-VLM with reactive replanning approaches such as VILA and RePlan-VLM under disturbance scenarios involving object addition and removal. we omit goal change as it requires explicit support for mid-execution instruction updates, which is not modeled in the original VILA and RePlan-VLM frameworks. In the object-addition setting, if the introduced new object blocks access to another object that must be manipulated based on the current task plan, that will lead to collision and  object damage in reactive methods. In contrast, ProAct-VLM detects the obstruction during execution and updates the task plan before attempting invalid actions. In object-removal scenarios such as S5, reactive approaches may attempt to manipulate a non-existent object before re-planning is invoked, reducing efficiency. Continuous monitoring enables ProAct-VLM to immediately revise the plan upon detection of the object's removal, avoiding unnecessary actions. Therefore, mid-execution disturbance detection improves robustness and reduces redundant motion compared to reactive  re-planning.

\subsection{Experiment setup   }
\begin{figure}[t]
    \centering
    \includegraphics[width=0.5\textwidth,height=5cm]{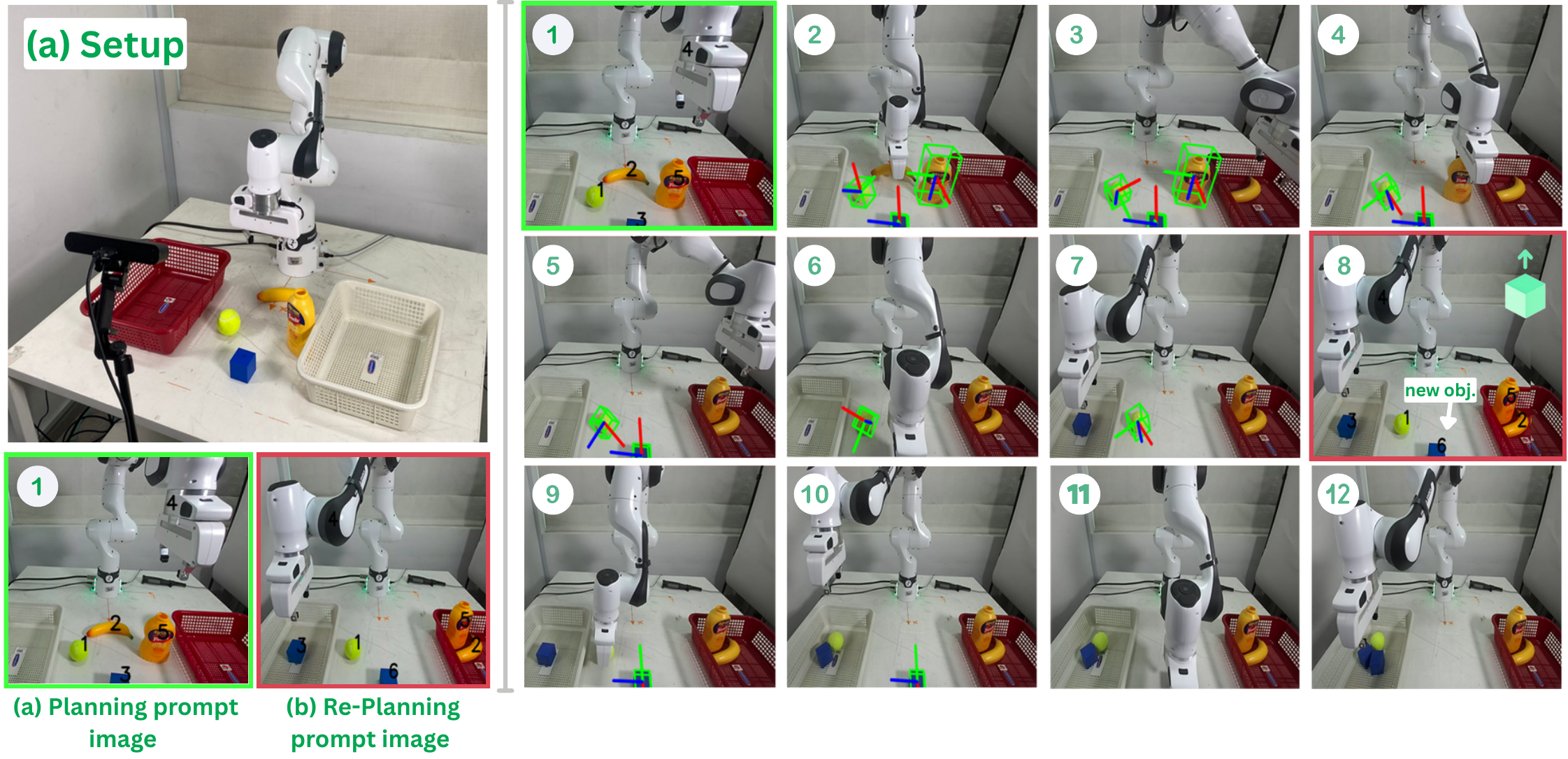}
    \caption{\small  The robot executes an initial VLM-generated plan; after a new object is introduced and detected by $\mathcal{E}$ (screenshot~8), the plan monitor triggers re-planning. The updated plan is executed successfully (screenshots~9--12).}
    \label{fig:experiment}
\end{figure}

To demonstrate the functionality of ProAct-VLM, we conducted a real-world execution trial of the sorting task described previously. The experimental setup is shown in Figure~\ref{fig:experiment}. Before execution, we performed camera-to-robot calibration to ensure accurate 6-DoF pose estimation of objects by $\mathcal{E}$ relative to the robot base frame. 
Initially, the augmented planning prompt image and task instructions were provided to the VLM to generate an initial plan, which the robot began executing. During execution, an additional object was introduced into the scene (screenshot~8). Which was detected and pose-estimated by $\mathcal{E}$, prompting the plan monitor to trigger a re-planning process. The VLM was then prompted with the updated augmented image and generated a revised plan, which the robot executed successfully (screenshots~9--12).

\section{conclusion}
In this paper, we introduce ProAct-VLM, a task planning framework that integrates VLMs with a real-time perception and feedback loop for robust, physically grounded robotic task planning. Unlike prior methods that rely on reactive or discrete pre-action checks, ProAct-VLM continuously monitors the environment and re-invokes the VLM whenever task-relevant changes that may invalidate the current plan are detected, enabling adaptation before failure occurs. We validated the framework on a real-world sorting task in which disturbances such as object addition, removal, and goal modification were introduced at different stages of execution. A comparative analysis against baselines confirmed that ProAct-VLM consistently improves both success rates and efficiency in dynamic, long-horizon manipulation tasks.
Our current implementation assumes a predefined task-relevant object vocabulary and relies on CAD-based pose estimation, making it best suited for environments where manipulable objects are known in advance. We also do not evaluate robustness under heavy perception noise or extreme clutter. Extending the framework toward open-set perception and uncertainty-aware disturbance modeling remains an important direction for future work. Moreover, we aim to extend ProAct-VLM beyond task-level adaptation to incorporate adaptive motion planning. By leveraging the continuously updated environment state, the framework can be integrated with feedback-driven motion policies that adjust trajectories online, enabling tighter coupling between perception, reasoning, and control.

\section{Acknowledgment}
This research was supported by the Center for Autonomous Robotic Systems (CARS), Khalifa University of Science and Technology (KU-CARS), through the project "T2FS (TactileThumbFirmSense) Device: A Wearable Thumb Device That is Capable of Sensing Fruit Firmness Using Vision-Based Tactile Sensors" by Silal, under Project ID: KU-EXT-SILAL-2025-8475000023.
\bibliographystyle{IEEEtran}
\bibliography{main}

@inproceedings{wang2024llm3,
  title={Llm3: Large language model-based task and motion planning with motion failure reasoning},
  author={Wang, Shu and Han, Muzhi and Jiao, Ziyuan and Zhang, Zeyu and Wu, Ying Nian and Zhu, Song-Chun and Liu, Hangxin},
  booktitle={2024 IEEE/RSJ International Conference on Intelligent Robots and Systems (IROS)},
  pages={12086--12092},
  year={2024},
  organization={IEEE}
}

@inproceedings{kirillov2023segment,
  title={Segment Anything},
  author={Kirillov, Alexander and Mintun, Eric and Ravi, Nikhila and Mao, Hanzi and Rolland, Chloe and Gustafson, Laura and Xiao, Tete and Whitehead, Spencer and Berg, Alexander C and Lo, Wan-Yen and others},
  booktitle={2023 IEEE/CVF International Conference on Computer Vision (ICCV)},
  pages={3992--4003},
  year={2023},
  organization={IEEE Computer Society}
}

@inproceedings{rana2023sayplan,
  author    = {Rana, Krishan and Haviland, Jason and Garg, Saurabh and Abou-Chakra, Jad and Reid, Ian and S{\"u}nderhauf, Niko},
  title     = {SayPlan: Grounding Large Language Models using 3D Scene Graphs for Scalable Robot Task Planning},
  booktitle = {Proceedings of the 7th Conference on Robot Learning (CoRL)},
  series    = {Proceedings of Machine Learning Research},
  year      = {2023},
  pages     = {23--72}
}

@inproceedings{liu2024grounding,
  title={Grounding dino: Marrying dino with grounded pre-training for open-set object detection},
  author={Liu, Shilong and Zeng, Zhaoyang and Ren, Tianhe and Li, Feng and Zhang, Hao and Yang, Jie and Jiang, Qing and Li, Chunyuan and Yang, Jianwei and Su, Hang and others},
  booktitle={European Conference on Computer Vision},
  pages={38--55},
  year={2024},
  organization={Springer}
}

@article{yang2022decoupling,
  title={Decoupling features in hierarchical propagation for video object segmentation},
  author={Yang, Zongxin and Yang, Yi},
  journal={Advances in Neural Information Processing Systems},
  volume={35},
  pages={36324--36336},
  year={2022}
}

@inproceedings{wen2024foundationpose,
  title={Foundationpose: Unified 6d pose estimation and tracking of novel objects},
  author={Wen, Bowen and Yang, Wei and Kautz, Jan and Birchfield, Stan},
  booktitle={Proceedings of the IEEE/CVF Conference on Computer Vision and Pattern Recognition},
  pages={17868--17879},
  year={2024}
}

@ARTICLE{vlm_replan,
  author={Mei, Aoran and Zhu, Guo-Niu and Zhang, Huaxiang and Gan, Zhongxue},
  journal={IEEE Robotics and Automation Letters}, 
  title={ReplanVLM: Replanning Robotic Tasks With Visual Language Models}, 
  year={2024},
  volume={9},
  number={11},
  pages={10201-10208},
  doi={10.1109/LRA.2024.3471457}}

@article{hu2023look,
  title={Look before you leap: Unveiling the power of gpt-4v in robotic vision-language planning},
  author={Hu, Yingdong and Lin, Fanqi and Zhang, Tong and Yi, Li and Gao, Yang},
  journal={arXiv preprint arXiv:2311.17842},
  year={2023}
}

@article{tziafas2024openworldgraspinglargevisionlanguage,
      title={Towards Open-World Grasping with Large Vision-Language Models}, 
      author={Georgios Tziafas and Hamidreza Kasaei},
      year={2024},
      journal={8th Conference on Robot Learning (CoRL 2024)},
}

@misc{ramachandran2025doesgpt4ounderstandvision,
  title        = {How Well Does GPT-4o Understand Vision? Evaluating Multimodal Foundation Models on Standard Computer Vision Tasks},
  author       = {Ramachandran, Rahul and Garjani, Ali and Bachmann, Roman and Atanov, Andrei and Kar, O\u{g}uzhan Fatih and Zamir, Amir},
  year         = {2025},
  eprint       = {2507.01955},
  archivePrefix= {arXiv},
  primaryClass = {cs.CV},
  url          = {https://arxiv.org/abs/2507.01955},
}

@article{a3,
  title   = {Inner Monologue: Embodied Reasoning Through Planning with Language Models},
  author  = {Huang, Wenlong and Fei, Jiajun and Mandlekar, Ajay and Gupta, Animesh},
  journal = {arXiv preprint arXiv:2207.05608},
  year    = {2022}
}

@inproceedings{zhao2023chat,
  title     = {Chat with the Environment: Interactive Multimodal Perception Using Large Language Models},
  author    = {Zhao, Xufeng and Lin, Yikai and Wu, Chengrui and Liu, Yiming and Zhang, Weicong and Wang, Hang},
  booktitle = {2023 IEEE/RSJ International Conference on Intelligent Robots and Systems (IROS)},
  year      = {2023},
  publisher = {IEEE}
}

@article{replan,
  title={RePLan: Robotic Replanning with Perception and Language Models},
  author={Zhao, Xiang and Xu, Xiaohan and Zhu, Yilun and Szot, Andrew and Xiao, Fei and Savarese, Silvio and Fei-Fei, Li and Yue, Yixin and Gao, Yixiao and Zhang, Yunzhu},
  journal={arXiv preprint arXiv:2401.04157},
  year={2024},
  url={https://arxiv.org/abs/2401.04157}
}

@inproceedings{wang2024llm,
  author    = {Wang, R. and Yang, Z. and Zhao, Z. and Tong, X. and Hong, Z. and Qian, K.},
  title     = {LLM-based Robot Task Planning with Exceptional Handling for General Purpose Service Robots},
  booktitle = {2024 43rd Chinese Control Conference (CCC)},
  year      = {2024},
  pages     = {4439--4444},
  publisher = {IEEE},
  month     = jul,
  doi       = {10.23919/CCC65962.2024.10428425}
}

@article{liu2025robodexvlm,
  title={Robodexvlm: Visual language model-enabled task planning and motion control for dexterous robot manipulation},
  author={Liu, Haichao and Guo, Sikai and Mai, Pengfei and Cao, Jiahang and Li, Haoang and Ma, Jun},
  journal={arXiv preprint arXiv:2503.01616},
  year={2025}
}

@book{wilensky1983planning,
  title={Planning and Understanding: A Computational Approach to Human Reasoning},
  author={Wilensky, Robert},
  year={1983},
  publisher={Addison-Wesley}
}

@inproceedings{kim2024pre,
  title={Pre-emptive Action Revision by Environmental Feedback for Embodied Instruction Following Agents},
  author={Kim, Jinyeon and Min, Cheolhong and Kim, Byeonghwi and Choi, Jonghyun},
  booktitle={8th Annual Conference on Robot Learning},
  year={2024}
}

\end{document}